\documentclass{article}
\usepackage{ijcai26}

\usepackage{times}
\usepackage{soul}
\usepackage{url}
\usepackage[hidelinks]{hyperref}
\usepackage[utf8]{inputenc}
\usepackage[small]{caption}
\usepackage{graphicx}
\usepackage{amsmath}
\usepackage{amsthm}
\usepackage{booktabs}
\usepackage[switch]{lineno}

\usepackage[ruled,linesnumbered]{algorithm2e}
\usepackage{multirow}
\usepackage{colortbl}
\usepackage{xcolor}
\usepackage{amsfonts}

\definecolor{degrade}{rgb}{1, 0.9, 0.9}

\definecolor{rowgray}{gray}{0.89}

\title{Safeguarding Facial Identity against Diffusion-based Face Swapping via Cascading Pathway Disruption}

\author{
Liqin Wang$^1$
\and
Qianyue Hu$^1$
\and
Wei Lu$^{1,}$\thanks{Corresponding author.}
\and
Xiangyang Luo$^2$\\
\affiliations
$^1$MoE Key Laboratory of Information Technology, Sun Yat-sen University, Guangzhou, China\\
$^2$ State Key Laboratory of Mathematical Engineering and Advanced Computing, Zhengzhou, China\\
\emails
wanglq37@mail2.sysu.edu.cn, huqy56@mail2.sysu.edu.cn, \\
luwei3@mail.sysu.edu.cn,
luoxy\_ieu@sina.com,
}

\begin{document}

\maketitle

\begin{abstract}
The rapid evolution of diffusion models has democratized face swapping but also raises concerns about privacy and identity security. Existing proactive defenses, often adapted from image editing attacks, prove ineffective in this context. We attribute this failure to an oversight of the structural resilience and the unique static conditional guidance mechanism inherent in face swapping systems. To address this, we propose VoidFace, a systemic defense method that views face swapping as a coupled identity pathway. By injecting perturbations at critical bottlenecks, VoidFace induces cascading disruption throughout the pipeline. Specifically, we first introduce localization disruption and identity erasure to degrade physical regression and semantic embeddings, thereby impairing the accurate modeling of the source face. We then intervene in the generative domain by decoupling attention mechanisms to sever identity injection, and corrupting intermediate diffusion features to prevent the reconstruction of source identity. To ensure visual imperceptibility, we perform adversarial search in the latent manifold, guided by a perceptual adaptive strategy to balance attack potency with image quality. Extensive experiments show that VoidFace outperforms existing defenses across various diffusion-based swapping models, while producing adversarial faces with superior visual quality.
\end{abstract}

\begin{figure}[t]
    \centering
    \includegraphics[width=\columnwidth]{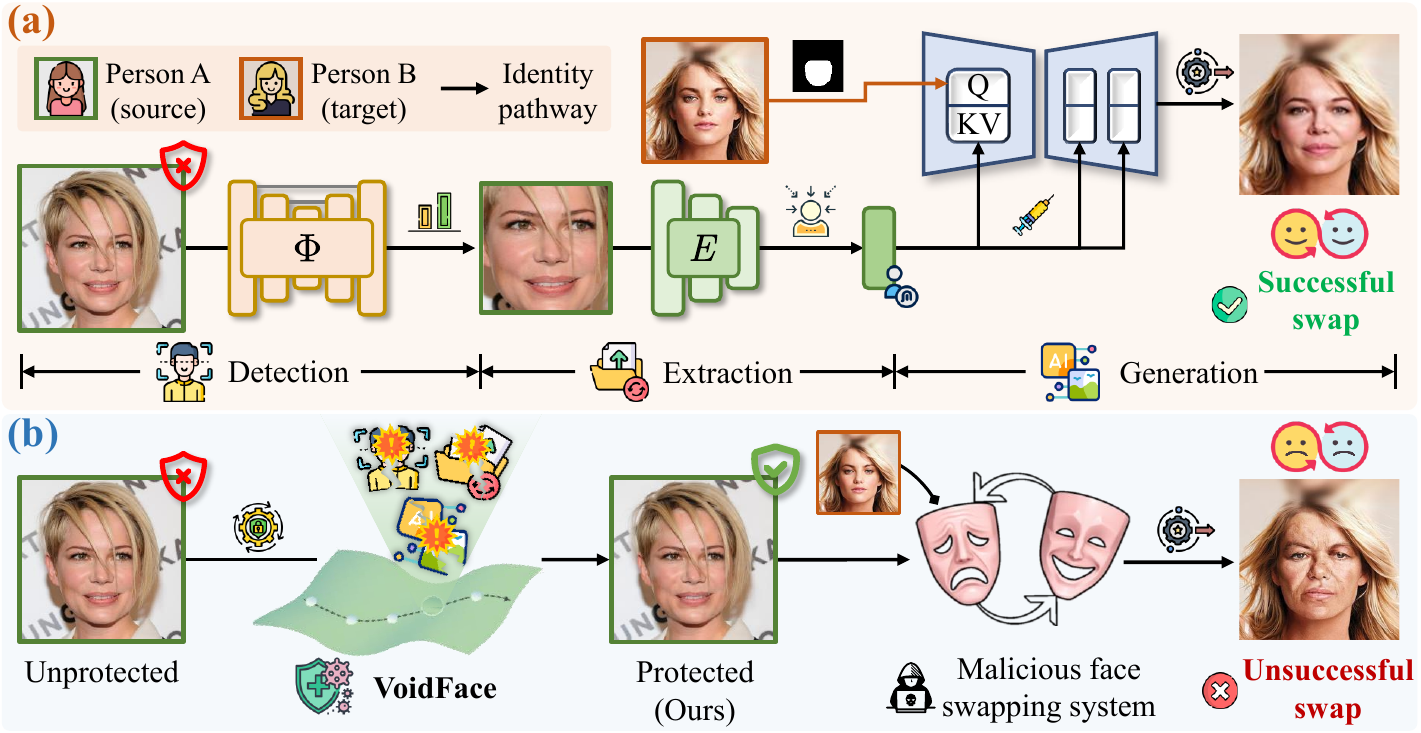}
    \caption{(a) Diffusion-based face swapping operates as a coupled identity pathway across multiple stages. (b) Our proposed VoidFace prevents unauthorized face swapping, safeguarding facial identity.}
    \label{fig:visual-1-intro_pic}
\end{figure}

\section{Introduction}
The rapid evolution of deep generative models has catalyzed significant advancements in Deepfake facial manipulation~\cite{deepfake_develop1,deepfake_develop2}, particularly in the area of face swapping. While this technology enables creative applications in entertainment and digital art, it has also been exploited for criminal activities, leading to the creation of facial privacy and societal problems. Consequently, there is an urgent need to develop effective defense mechanisms.

In response to these threats, existing countermeasures can be broadly categorized into passive detection and proactive defense. While passive detection~\cite{passive_detection1,passive_detection3} aims to distinguish synthetic content, it fails to interdict forgery at the source. In contrast, proactive defense offers a fundamental solution by embedding imperceptible perturbations to immunize face images against unauthorized manipulation. However, most previous research~\cite{proactive_gan1,proactive_gan2} has concentrated on GAN-based architectures. They overlook the unique mechanisms and internal structures of diffusion models~\cite{ddpm,ddim,ldm}, which limits their efficacy against emerging diffusion-based face swapping.

Although adversarial attacks against diffusion-based image editing exist, they exhibit a fundamental incompatibility with face swapping. First, unlike general editing, face swapping operates as a complex system with significant structural resilience (Fig. ~\ref{fig:visual-1-intro_pic} (a)). The strong generative priors and complex designs of these models allow them to correct intermediate errors. Consequently, previous attacks targeting isolated components (e.g., identity encoders) fail to disrupt the generation trajectory, as their perturbations are mitigated by the error-correction mechanisms. Second, current methods overlook a critical paradigm misalignment. While they typically perturb the noisy latent, face swapping utilizes the source image as a static semantic condition to guide synthesis. Failing to sever this specific conditional guidance renders existing attacks ineffective, as the source identity continues to steer the generation. Finally, existing methods typically apply global perturbations in the pixel space, which introduces visible artifacts and exhibits vulnerability to common pre-processing operations like resizing and compression.

To address these challenges, we propose VoidFace. As illustrated in Fig. \ref{fig:visual-1-intro_pic}, modern face swapping systems generally follow a three-stage workflow, similar to conventional Deepfakes: detection, extraction, and generation. We view this workflow not as disjoint modules, but as a tightly coupled identity pathway, where the fidelity of each stage strictly depends on its predecessor. Based on this insight, our method pivots from isolated attacks to systemic disruption. We exploit this sequential dependence as a primary vulnerability. By embedding perturbations at critical bottlenecks, we ensure that minor upstream errors are amplified downstream, ultimately leading to identity pathway cascading disruption.

Specifically, we decompose the adversarial objective into three progressive stages to ensure the destruction of identity information. First, addressing errors in the physical domain, we recognize that accurate face localization is fundamental to feature extraction. Thus, we introduce localization disruption to misalign the facial region, causing an initial loss of facial information. Second, building upon this destabilized baseline, we execute identity erasure in the semantic domain. We design a contrastive objective to drive the source representation toward a non-informative state. This not only obstructs the comprehensive modeling of facial semantics but also reduces the quantity of identity information accessible to the generative model. Third, to prevent the diffusion model from utilizing generative priors for reconstruction, we propose a targeted strategy to truncate the internal links in the generative domain. We introduce attention decoupling to interfere with the injection of the identity condition, preventing the attention mechanism from correctly aligning the source identity with the target structure. Simultaneously, acknowledging that the model relies on intermediate U-Net~\cite{unet} features to reconstruct and maintain identity, we design feature corruption to prevent the model from identifying and refining key facial areas. Through this interconnected strategy, we foster a cumulative error effect throughout the entire pipeline, thereby inhibiting unauthorized face swapping.

To implement this strategy while ensuring visual imperceptibility, we diverge from pixel-level optimization. Instead, we formulate the adversarial search within the low-dimensional latent manifold of a diffusion model, leveraging its effective generative prior to produce natural and robust adversarial faces. Furthermore, we introduce a perceptual adaptive strategy. By evaluating visual perception sensitivity to dynamically modulate perturbation strength, we achieve a superior balance between attack potency and visual quality.

To summarize, our main contributions are as follows:

\begin{itemize}    
    \item We propose VoidFace, a systemic defense method tailored for diffusion-based face swapping. By exploiting the sequential dependence of the identity pathway, we engineer a cascading disruption strategy that accumulates and amplifies errors throughout the pathway.

    \item We devise a progressive adversarial objective to degrade identity integrity. We induce localization disruption and identity erasure to weaken identity representation, while implementing attention decoupling and feature corruption to prevent identity integration in the generative domain, significantly inhibiting identity replication.

    \item We conduct adversarial optimization in the diffusion latent manifold with perceptual adaptation. This mechanism leverages visual feedback to guide the perturbation, achieving robust defense and visual imperceptibility.
\end{itemize}

\section{Related Work}
\textbf{Diffusion-based face swapping.}
With advancements in generative models, face swapping technology has evolved from GAN-based methods to diffusion-based approaches. Despite architectural shifts, diffusion-based face swapping generally follows a three-stage workflow: face detection, feature extraction, and face generation. These methods generally employ detectors like MTCNN~\cite{mtcnn} and RetinaFace~\cite{retinaface} for face localization and landmark alignment. They also utilize the pre-trained face model ArcFace~\cite{arcface} to extract identity embeddings. These embeddings are then injected into the diffusion process to condition the generation. Early approaches like DiffFace~\cite{diffface} and DiffSwap~\cite{diffswap} formulate the task as conditional reconstruction or inpainting, which requires model retraining with auxiliary constraints. Recently, adapter-based~\cite{ip-adapter} methods like Face-Adapter~\cite{faceadapter} and InstantID~\cite{instantid}, leverage lightweight modules to inject identity features into pre-trained backbones, enabling identity transfer without extensive retraining. The accessibility of these tools exacerbates privacy risks, underscoring the need for protective measures.

\paragraph{Diffusion adversarial attack.}
As diffusion models gain popularity in image editing, adversarial attacks against these architectures have progressed significantly. For instance, AdvDM~\cite{advdm} utilized the model’s inherent representations for attacks. PhotoGuard~\cite{photoguard} attacks the encoder to distort the latent representation of input images. Mist~\cite{mist} improves transferability by integrating semantic loss~\cite{advdm} and textural loss~\cite{photoguard}. Recently, methods like FACELOCK~\cite{facelock} and FaceShield~\cite{faceshield} focus on impeding facial feature extraction for identity protection.
However, these strategies primarily safeguard training data or query inputs, failing to account for the structural resilience and conditional guidance unique to face swapping. The paradigm misalignment limits their effectiveness to sever the link between the source identity and the output. Moreover, they typically apply pixel-level perturbations, which results in visible artifacts and fragility against pre-processing. To bridge these gaps, we propose a defense tailored for diffusion-based face swapping.
\begin{figure*}[t]
    \centering
    \includegraphics[width=0.914\textwidth]{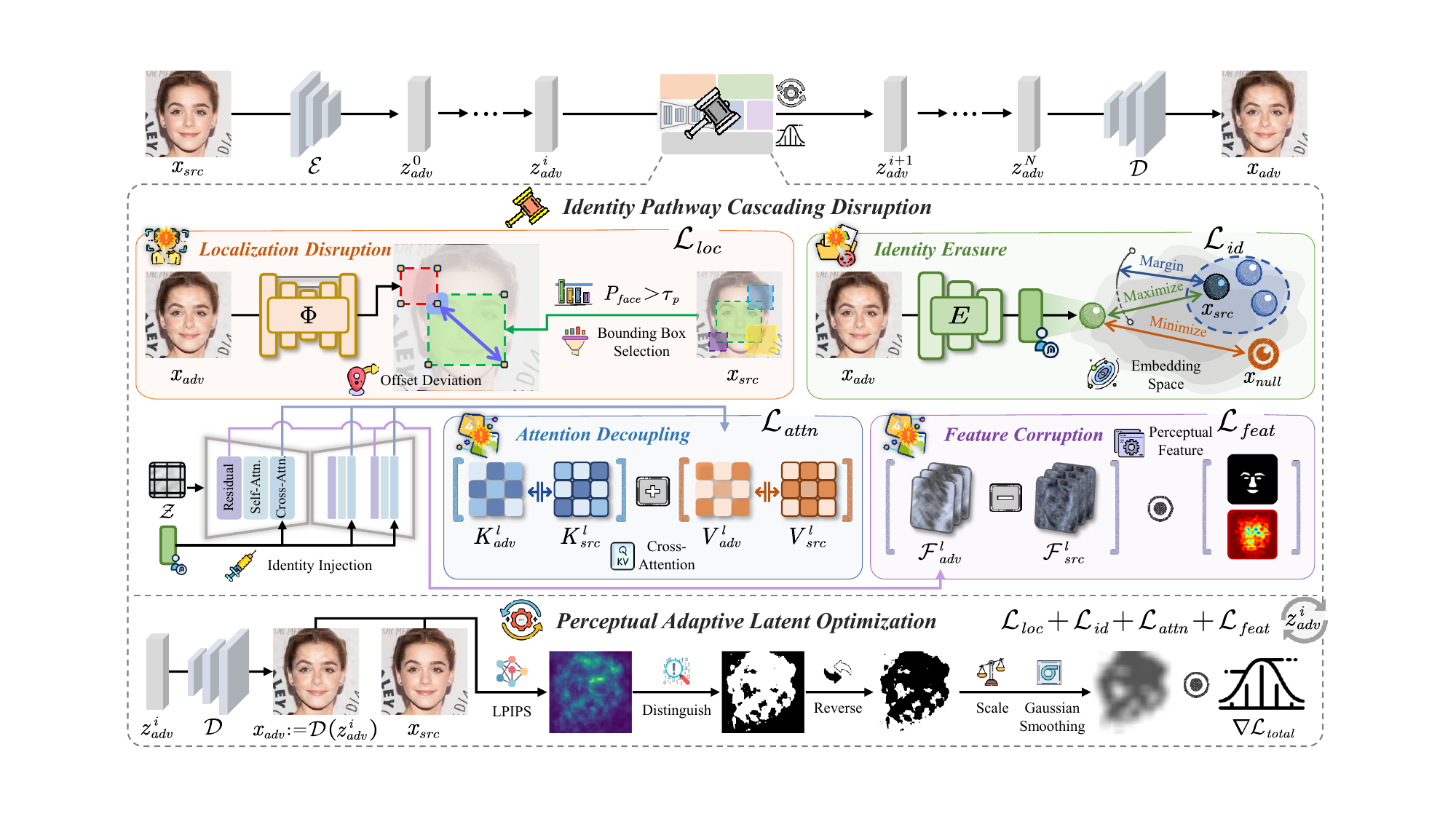}
    \caption{Overview of the proposed VoidFace. We formulate the adversarial defense as a cascading disruption of the identity pathway. We systematically target critical bottlenecks across the physical, semantic, and generative domains. The perceptual adaptive latent optimization strategy is integrated to adaptively balance attack potency and visual quality of adversarial faces.}
    \label{fig:visual-8-overview}
\end{figure*}

\section{Method}
\subsection{Overview}
Given a source face image $x_{src}$,  we aim to generate an adversarial face $x_{adv}$ that safeguards facial identity against unauthorized diffusion-based face swapping. As shown in Fig .~\ref{fig:visual-1-intro_pic}, modern face swapping systems typically operate as a three-stage workflow. First, in the detection stage, the system localizes the precise facial region to align and standardize the input. The aligned face is then encoded into compact semantic embeddings that capture identity features in the extraction stage. For generation, these embeddings are injected as condition to guide the model to reconstruct facial details. This sequential dependence constitutes a tightly coupled identity pathway. Therefore, we propose VoidFace (shown in Fig. ~\ref{fig:visual-8-overview}) to systematically disrupt this pathway at critical bottleneck.

Our approach begins with localization disruption in the physical domain, which targets the bounding box regression to induce invalid localization, thereby destabilizing the detection stage. Next, we degrade the semantic domain during extraction via identity erasure, a contrastive objective that utilizes ensemble identity encoders to suppress identity information. For the generative domain, we introduce two interventions. First, attention decoupling perturbs the key and value matrices within the cross-attention layers, interfering with the conditional identity injection. Second, feature corruption targets intermediate features within the U-Net, preventing the model from identifying and reconstructing facial details. By exploiting the sequential dependence across stages, the induced perturbations accumulate and amplify, leading to a cascading disruption of the identity pathway. 

To ensure robustness and visual imperceptible, we perform the adversarial optimization within a low-dimensional latent manifold. We further propose a perceptual adaptive strategy to reduce artifacts in smooth facial regions. This strategy leverages a perception network to distinguish sensitive regions and adaptively adjusts perturbation strength, thereby balancing attack effectiveness and visual quality.

\subsection{Identity Pathway Cascading Disruption}
\label{sec: Disrupting Identity Conditional Pathways}
\paragraph{Localization disruption.}
Face swapping pipelines universally necessitate precise localization to align and standardize the input for feature extraction. We exploit this dependency to induce incorrect localization, compelling the downstream module to process the irrelevant background region instead of the facial area. 
Specifically, we attack the bounding box regression branch of the face detector $\Phi$. For a given input image $x$, the detector $\Phi$ predicts classification scores and regression offsets for all $J$ pre-defined anchors. We denote the classification output as a probability distribution $P_{face} \in [0, 1]^J$, representing the confidence of face existence. Correspondingly, the regression output $\Phi_{reg}(x) \in \mathbb{R}^{J \times 4}$ parameterizes $[\Delta_x, \Delta_y, \Delta_w, \Delta_h]$, which encode the spatial shifts and scaling factors required to refine the default anchors into precise bounding boxes. 
To concentrate the attack on valid face candidates, we propose a masking technique to retain anchors where the face confidence exceeds a threshold $\tau_p$
\begin{equation}
\mathcal{M}_{p} = \mathbb{I}[P_{face}(x_{src}) > \tau_p]
\end{equation}
where $\mathbb{I}$ is the indicator function. We then formulate the localization disruption loss by maximizing the offset deviation
\begin{equation}
\mathcal{L}_{loc} = \exp \left( - \| (\Phi_{reg}(x_{adv}) - \Phi_{reg}(x_{src})) \odot \mathcal{M}_{p} \|_2 \right)
\end{equation}
where $\odot$ denotes the Hadamard product. By minimizing $\mathcal{L}_{loc}$, we push the predicted bounding boxes away from their desired locations in the physical domain, causing an initial alignment failure for the subsequent stages.

\paragraph{Identity erasure.}
Following the physical disruption, we proceed to the semantic domain to disrupt the accurate extraction of source identity information. We target the identity encoder $E$, which compresses facial semantics into compact embedding vectors. We design a contrastive objective to suppress identity information by shifting the adversarial embedding toward a non-informative anchor $x_{null}$ (e.g., a black image) while pushing it away from the source identity. To enhance transferability, we ensemble the most commonly used pre-trained backbones~\cite{arcface} as the target encoders. Consequently, the identity erasure loss is defined as
\begin{equation}
\begin{split}
    \mathcal{L}_{id} = \, & D_{cos}(E(x_{adv}), E(x_{null})) \\
    & + \max\left(0, m - D_{cos}(E(x_{adv}), E(x_{src}))\right)
\end{split}
\end{equation}
where $D_{cos}(u, v) = 1 - \frac{u \cdot v}{\|u\|_2\|v\|_2}$ denotes the cosine distance and $m$ is the margin. Minimizing $\mathcal{L}_{id}$ disrupts the semantic content of the source face, depriving the downstream generative process of valid identity information.

\paragraph{Attention decoupling.}
Entering the generative domain, we interfere with the injection of identity information. We target the cross-attention mechanism, which is the key to integrating identity semantics into the denoising U-Net. Structurally, the noisy latent $\mathcal{Z}_{t}$ is projected into query $Q$, determining the spatial layout and pose of the generated face. The source identity embedding serves as the condition, yielding $\mathcal{C}_{id}=E(x)$, which is projected into key $K$ and value $V$ to guide synthesis. For a specific layer $l$, the attention mechanism is defined as
\begin{align}
    &~~ Q^l = \ell_q^l(\mathcal{Z}_{t}), ~~ K^l=\ell_k^l(\mathcal{C}_{id}), ~~ V^l=\ell_v^l(\mathcal{C}_{id}), \\
    & \text{Attention}(Q, K, V) = \text{Softmax}  \left(\frac{Q^l (K^l)^T}{\sqrt{d}} \right) V^l
\end{align}
where $\ell_{(\cdot)}^l$ denotes the linear projections and $d$ is the feature dimension. Crucially, the key matrix interacts with query to align identity features with the target spatial structure, while the value matrix determines the specific identity semantics to be injected. To disrupt this mechanism, we explicitly perturb the key and value matrices. Let $K_{adv}^l, V_{adv}^l$ and $K_{src}^l$, $V_{src}^l$ denote the matrices derived from the adversarial and source images, respectively. We formulate the attention decoupling loss by maximizing the difference between these matrices
\begin{equation}
\begin{split}
\mathcal{L}_{attn} = \sum_{l \in \Omega} \big( \| K_{adv}^l - K_{src}^l \|_2 
& + \| V_{adv}^l - V_{src}^l \|_2 \big)
\end{split}
\end{equation}
where $\Omega$ denotes the set of cross-attention layers. By maximizing $\mathcal{L}_{attn}$, we simultaneously misalign the spatial correspondence between the target structure and the identity condition, and degrade the semantic fidelity of the injected features. This effectively decouples the source identity from the generation trajectory.

\paragraph{Feature corruption.}
Simultaneously, we implement the second intervention within the generative domain. Existing research establishes that perceptual features within the U-Net form the foundation for model recognition and generation~\cite{feature_perceptual}. In face swapping, perceptual features reflect the model's understanding of facial structure and texture, and the model relies on these perceptual features to reconstruct and maintain identity. Therefore, our objective is to corrupt these features, preventing the source identity from being successfully reconstructed and transferred.

Specifically, let $\mathcal{F}_{adv}^l$ and $\mathcal{F}_{src}^l$ denote the feature maps extracted from layer $l$ of U-Net using the adversarial image $x_{adv}$ and source image $x_{src}$ as conditions, respectively. We target the first down-sampling block $l_{down}$ and the last up-sampling block $l_{up}$ of the U-Net, as these high-resolution layers are important for synthesizing detailed facial areas.

To confine the distortion to identity-aware areas, we integrate two masks: i) a semantic mask $\mathcal{M}_{sem}$ generated by a face parsing network~\cite{bisenet}. This segments biometric components (e.g., eyes, nose, and mouth) that are vital for human visual recognition. ii) a heatmap mask $\mathcal{M}_{cam}$ derived via Layer-CAM~\cite{layercam} on the identity encoder. This highlights the activation regions prioritized by recognition models. Both masks are downsampled to match the feature resolution. Formally, the feature corruption loss is
\begin{equation}
\mathcal{L}_{feat} =  \sum_{l \in \mathcal{S}} \sum_{k \in \mathcal{K}} \left\| \left( \mathcal{F}_{adv}^l - \mathcal{F}_{src}^l \right) \odot \mathcal{M}_k \right\|_2
\end{equation}
where $\mathcal{S}=\{l_{down},l_{up}\}$ and $\mathcal{K}=\{sem,cam\}$. Maximizing $\mathcal{L}_{feat}$ disrupts the identity-aware areas in perceptual features, thereby weakening the ability of the model to identify and reconstruct key identity details from the source image.

\paragraph{Total loss.}
Accordingly, the total loss function is defined as
\begin{equation}
    \mathcal{L}_{total} = (\mathcal{L}_{loc}, 
    \mathcal{L}_{id},
    \mathcal{L}_{attn}, \mathcal{L}_{feat})\Lambda^T
\end{equation}
where $\Lambda=(\lambda_{loc}, \lambda_{id}, \lambda_{attn}, \lambda_{feat})$ controls the strength of the respective loss term. Note that $\lambda_{loc}$ and $\lambda_{id}$ are negative, while $\lambda_{attn}$ and $\lambda_{feat}$ are positive in our setting.

\subsection{Perceptual Adaptive Latent Optimization}
\label{sec: Perceptual Adaptive Optimization}
\paragraph{Latent space optimization.}
To mitigate artifacts and vulnerability caused by pixel-level optimization, we conduct adversarial search in the low-dimensional latent manifold of a diffusion model. This manifold represents many contents of natural images, allowing the generated faces to maintain high visual quality and contain richer and more robust adversarial contents (e.g., texture and color)~\cite{diffaim}.

Specifically, we utilize the encoder $\mathcal{E}$ of VAE to map the source image $x_{src}$ into the latent space, initializing the adversarial latent code as $z_{adv}^0 = \mathcal{E}(x_{src})$. 
At each iteration $i$, we generate a temporary adversarial face using the VAE decoder, yielding $x_{adv}^i = \mathcal{D}(z_{adv}^{i})$. This adversarial face is then used to compute the total loss function. We restrict the adversarial search close to the clean image manifold learned by the generative model, which can be expressed as
\begin{equation}
z_{adv}^{i+1} = z_{adv}^{i} + \alpha \cdot \text{sign}(\nabla_{z_{adv}^{i}} \mathcal{L}_{total})
\label{eq:update_rule}
\end{equation}
where $\alpha$ is the step size. After $N$ iterations, we obtain the final adversarial face image $x_{adv} = \mathcal{D}(z_{adv}^{N})$, subject to the constraint that $\|x_{adv}-x_{src}\|_{\infty} \le \epsilon$.

\begin{figure*}[t]
    \centering
    \includegraphics[width=0.9\textwidth]{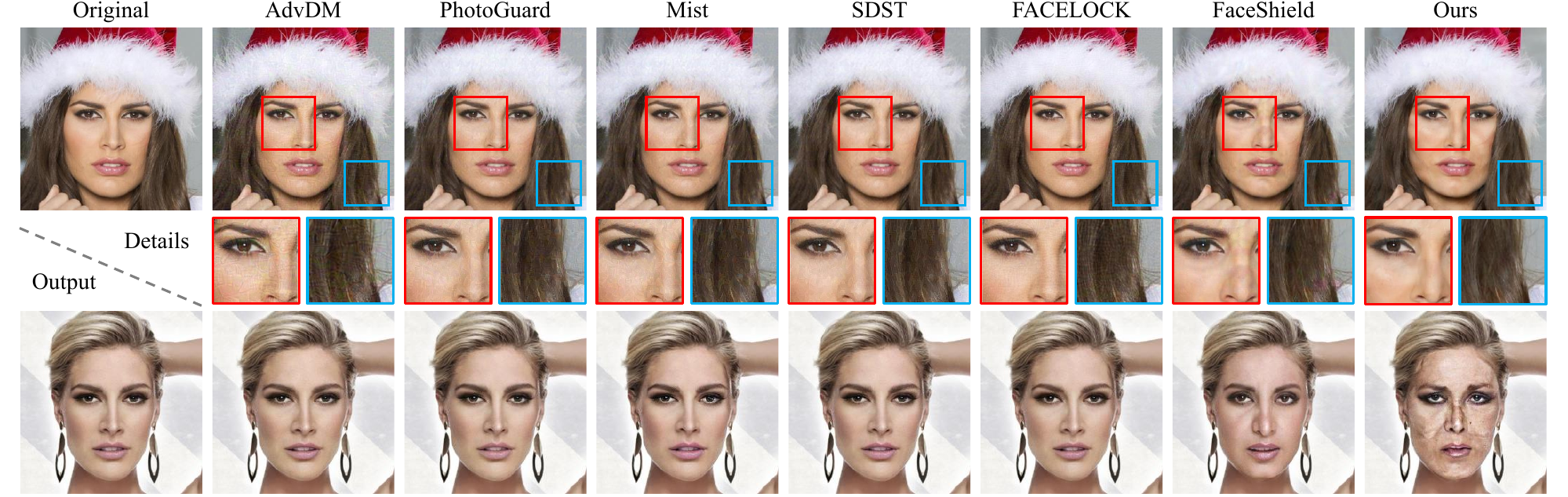}
    \caption{Comparison of visual quality and defense effectiveness against DiffFace. The top and middle rows display protected images and details. The baselines often introduce noticeable noise, while our method generates natural faces. The bottom row shows the face swapping outputs using the protected images as source face. We effectively disrupt face swapping, leading to severe artifacts and mismatched identity.}
    \label{fig:visual-2-intro}
\end{figure*}
\paragraph{Perceptual adaptive strategy.}
According to the Weber-Fechner law, humans are more sensitive to subtle changes in smooth regions than in textured ones. While latent space optimization ensures overall naturalness, it offers limited control over local perceptual effect. Due to the non-linear mapping between latent space and pixel space, uniform latent updates often cause perturbations to manifest as uneven artifacts in smooth facial regions. To address this, we introduce a strategy that spatially adjusts latent perturbation strength using perceptual feedback, reducing artifacts in sensitive regions.

Specifically, at each iteration $i$, we first compute the current LPIPS~\cite{lpips} distance map and reverse it to obtain a spatial perceptual map
\begin{equation}
\mathcal{S}^{i} = 1-\text{LPIPS}(\mathcal{D}(z_{adv}^{i}), x_{src})
\end{equation}
here, a higher value in $\mathcal{S}^{i}$ indicates a region with high visual fidelity, permitting stronger perturbations, whereas a lower value signals potential artifacts. We then generate a binary mask $\mathcal{M}^i$ to spatially distinguish these regions
\begin{equation}
    \mathcal{M}^i = \mathbb{I}[\mathcal{S}^{i} > \text{Quantile}(\mathcal{S}^{i}, q) ]   
\end{equation}
where $\text{Quantile}(\cdot, q)$ returns the $q$-th percentile threshold. Within $\mathcal{M}^i$, 1 signifies the least noticeable regions suitable for perturbation, and 0 indicates sensitive regions. However, directly applying $\mathcal{M}^i$ to constrain update introduces sharp transitions and leaves sensitive regions entirely unprotected, which reduces the overall protection efficacy. Thus, we further construct a smooth perceptual map
\begin{equation}
\mathcal{P}^{i} = \mathcal{G}_\sigma ( \mathcal{M}^{i} + \gamma \cdot (1 - \mathcal{M}^{i}))
\end{equation}
where $\gamma \in [0, 1]$ ensures sensitive regions take weakened perturbations to maintain necessary adversarial efficacy, and the Gaussian smoothing $\mathcal{G}_{\sigma}$ reduces the sharpening of transitions. We then downsample $\mathcal{P}^{i}$ to match the resolution of the latent code. Finally, the update rule in Eq. \ref{eq:update_rule} is reformulated as
\begin{equation}
z_{adv}^{i+1} = z_{adv}^{i} + \alpha \cdot \mathcal{P}^i \odot \text{sign}(\nabla_{z_{adv}^{i}} \mathcal{L}_{total})
\end{equation}

This strategy ensures the perturbation strength is locally and adaptively constrained by perceptual quality, achieving a balance between attack efficacy and imperceptibility.

\begin{figure*}[t]
    \centering
    \includegraphics[width=0.9\textwidth]{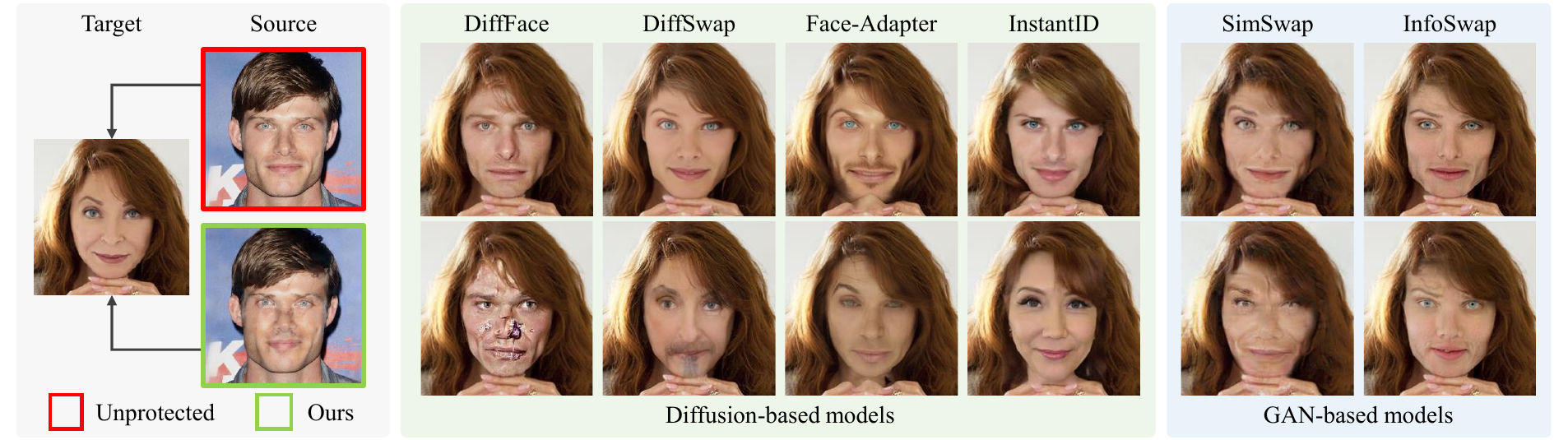}
    \caption{Qualitative evaluation of our protection efficacy. The evaluation covers a wide range of architectures, including four diffusion-based and two GAN-based models. As observed, our method effectively safeguards the source images, where the protected outputs suffer from severe visual degradation or complete identity mismatch, verifying both the effectiveness and transferability of our approach.}
    \label{fig:visual-main}
\end{figure*}
\begin{table*}[!t]
    \centering
    \resizebox{0.92\textwidth}{!}{
    \begin{tabular}{l ccl ccl ccl ccl} 
    \toprule
    \textbf{Model} & \multicolumn{3}{c}{DiffFace} & \multicolumn{3}{c}{DiffSwap} & \multicolumn{3}{c}{Face-Adapter} & \multicolumn{3}{c}{InstantID} \\
    \cmidrule(r){2-4} \cmidrule(lr){5-7} \cmidrule(lr){8-10} \cmidrule(l){11-13}
    \textbf{Metric} & $L_2 \uparrow$ & ISM $\downarrow$ & PSNR $\downarrow$ & $L_2 \uparrow$ & ISM $\downarrow$ & PSNR $\downarrow$ & $L_2 \uparrow$ & ISM $\downarrow$ & PSNR $\downarrow$ & $L_2 \uparrow$ & ISM $\downarrow$ & PSNR $\downarrow$ \\
    \midrule
    \textbf{Dataset} & \multicolumn{12}{c}{CelebA-HQ} \\
    \midrule
    AdvDM       & 0.0169 & 0.4115 & 35.7688 & 0.0411 & 0.2531 & 27.7711 & 0.0174 & 0.3838 & 35.5068 & 0.0629 & 0.3636 & 23.9884 \\
    PhotoGuard  & 0.0160 & 0.4126 & 36.2302 & 0.0409 & 0.2511 & 27.8350 & 0.0141 & 0.3912 & 37.4049 & 0.0622 & 0.3717 & 24.0949 \\
    Mist        & 0.0169 & 0.4143 & 35.8107 & 0.0406 & 0.2500 & 27.8813 & 0.0139 & 0.3899 & 37.5348 & 0.0615 & 0.3732 & 24.2018 \\
    SDST        & 0.0158 & 0.4137 & 36.3371 & 0.0409 & 0.2507 & 27.8271 & 0.0138 & 0.3933 & 37.6854 & 0.0615 & 0.3738 & 24.1969 \\
    FACELOCK    & 0.0151 & 0.4118 & 36.8193 & 0.0405 & 0.2502 & 27.8913 & 0.0129 & 0.3910 & 38.2923 & 0.0619 & 0.3713 & 24.1581 \\
    FaceShield  & 0.0351 & 0.3385 & 29.1058 & 0.0407 & 0.2496 & 27.8671 & 0.0181 & 0.3782 & 35.0284 & 0.0624 & 0.3617 & 24.0818 \\
    \midrule
    \rowcolor{rowgray}
    \textbf{Ours} & \textbf{0.0425} & \textbf{0.3256} & \textbf{27.4614} & \textbf{0.0416} & \textbf{0.2488} & \textbf{27.7005} & \textbf{0.0303} & \textbf{0.3438} & \textbf{30.4762} & \textbf{0.0672} & \textbf{0.2585} & \textbf{23.4100} \\
    \midrule
    \midrule
    \textbf{Dataset} & \multicolumn{12}{c}{VGGFace2-HQ} \\
    \midrule
    AdvDM       & 0.0274 & 0.3227 & 31.7209 & 0.0483 & 0.1851 & 26.3721 & 0.0299 & 0.3521 & 31.0866 & 0.0773 & 0.3283 & 22.1852 \\
    PhotoGuard  & 0.0291 & 0.3172 & 31.2410 & 0.0479 & 0.1847 & 26.4259 & 0.0263 & 0.3577 & 32.2220 & 0.0776 & 0.3315 & 22.1600 \\
    Mist        & 0.0279 & 0.3202 & 31.5062 & 0.0474 & 0.1839 & 26.5075 & 0.0261 & 0.3570 & 32.3322 & 0.0772 & 0.3345 & 22.2009 \\
    SDST        & 0.0280 & 0.3206 & 31.5794 & 0.0472 & 0.1858 & 26.5529 & 0.0248 & 0.3594 & 32.8109 & 0.0771 & 0.3351 & 22.2112 \\
    FACELOCK    & 0.0273 & 0.3218 & 31.7532 & 0.0472 & 0.1839 & 26.5586 & 0.0208 & 0.3612 & 34.5242 & 0.0775 & 0.3353 & 22.1699 \\
    FaceShield  & 0.0449 & 0.2721 & 27.0080 & 0.0476 & 0.1826 & 26.5016 & 0.0279 & 0.3476 & 31.5714 & 0.0783 & 0.3204 & 22.0882 \\
    \midrule
    \rowcolor{rowgray}
    \textbf{Ours} & \textbf{0.0502} & \textbf{0.2613} & \textbf{26.0250} & \textbf{0.0489} & \textbf{0.1812} & \textbf{26.2622} & \textbf{0.0407} & \textbf{0.3177} & \textbf{28.0224} & \textbf{0.0830} & \textbf{0.1959} & \textbf{21.5624} \\
    \bottomrule
    \end{tabular}
    }
    \caption{Quantitative comparison of protection effectiveness among baseline methods on four diffusion-based face swapping models using CelebA-HQ and VGGFace2-HQ datasets. \textbf{Bold} indicates the best results. We consistently achieve superior protection performance.}
    \label{table:comparison}
\end{table*}

%
\begin{table}[!t]
    \centering
    \setlength{\tabcolsep}{4pt}
    \renewcommand{\arraystretch}{1.1}
    \resizebox{\columnwidth}{!}{
    \begin{tabular}{l ccc c ccc} 
    \toprule
    & \multicolumn{3}{c}{CelebA-HQ} && \multicolumn{3}{c}{VGGFace2-HQ} \\
    \cmidrule{2-4} \cmidrule{6-8}
    \textbf{Method} & LPIPS $\downarrow$ & PSNR $\uparrow$ & FID $\downarrow$ && LPIPS $\downarrow$ & PSNR $\uparrow$ & FID $\downarrow$ \\
    \midrule
    AdvDM      & 0.4407 & 30.2482 & 80.9156 && 0.4518 & 30.2226 & 86.3086 \\
    PhotoGuard & 0.5685 & 29.7640 & 48.1924 && 0.5655 & 29.7787 & 52.3568 \\
    Mist       & 0.5667 & 29.8417 & 48.3273 && 0.5641 & 29.8653 & 53.3954 \\ 
    SDST       & 0.5533 & 31.2611 & 41.8947 && 0.5473 & 31.3101 & 45.1694 \\
    FACELOCK   & 0.4040 & 30.5176 & 39.0089 && 0.4286 & 30.4588 & 42.9250 \\
    FaceShield & 0.2069 & 31.8288 & 34.5463 && 0.2115 & 31.3455 & 37.5160 \\
    \midrule
    \rowcolor{rowgray}
    \textbf{Ours} & \textbf{0.1628} & \textbf{32.9606} & \textbf{32.5384} && \textbf{0.1688} & \textbf{31.8421} & \textbf{33.2325} \\
    \bottomrule
    \end{tabular}
    }
    \caption{Comparison of the image quality of adversarial faces.}
    
    \label{table:visual_quality}
\end{table}
\begin{figure}[t]
    \centering
    \includegraphics[width=\columnwidth]{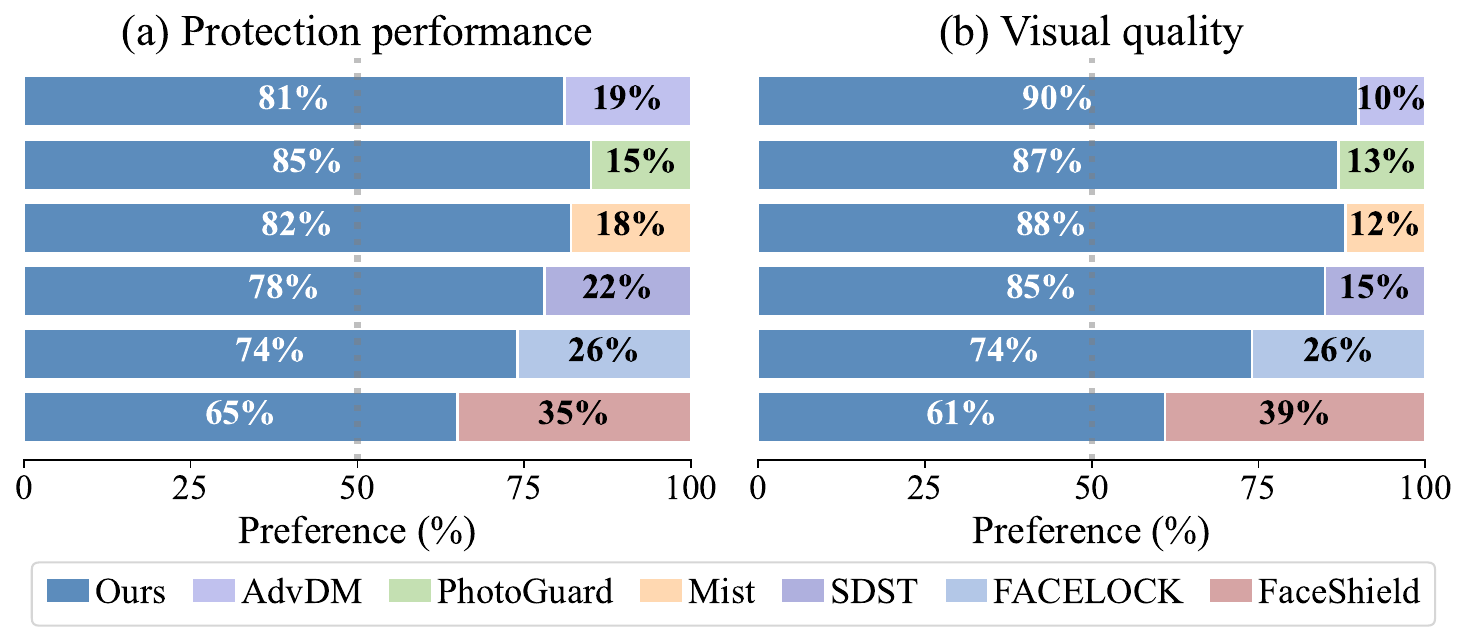}
    \caption{User preference study. Blue bars indicate the percentage of votes favoring our method.}
    \label{fig:visual-5-user_study}
\end{figure}

\section{Experiments}
\subsection{Setup}
\paragraph{Datasets.}
We validate our method on two datasets: CelebA-HQ~\cite{celebahq} and VGGFace2-HQ~\cite{vggface2hq}. The former contains 30,000 high-resolution celebrity face images, and the latter is a high-resolution version of VGGFace2 with 3.3 million face images from 9,131 unique identities. We randomly select 200 identities from each dataset, using 100 images for the source and 100 images for the target.

\paragraph{Baselines.}
We utilize four diffusion-based face swapping models as victim models to attack, including DiffFace, DiffSwap, Face-Adapter, and InstantID. Additionally, we investigate the cross-architecture transferability against GAN-based approaches in Sec. \ref{sec: transferability}. For comparative analysis, we benchmark against: AdvDM, PhotoGuard, Mist, SDST, FACELOCK, and Faceshield. All methods are evaluated under identical hyperparameter settings to ensure fair comparison.

\paragraph{Metrics.}
We assess protection effectiveness using $L_2$, Identity Score Matching~\cite{ism} (ISM), and PSNR. The $L_2$ and PSNR metrics evaluate image quality by comparing face swapping results of clean and protected images, with higher $L_2$ and lower PSNR indicating more distortion. ISM measures the similarity between the source face and the face swapping output, with lower values indicating less similarity. We also analyze the visual quality of adversarial face images using LPIPS, PSNR, and FID~\cite{fid}.

\paragraph{Implementation details.}
We generate adversarial faces by adopting Stable Diffusion v1.5 as the pre-trained diffusion model. We employ MTCNN and RetinaFace as the face detectors, and an ensemble of ArcFace variants as the identity encoders. We set the perturbation budget $\epsilon=12/255$, the step size $\alpha=1/255$, and total iterations $N=30$. Further details are available in the Supplementary Material.

\subsection{Comparison Study}
\paragraph{Protection performance.}
We quantitatively compare our method with state-of-the-art baselines. As reported in Tab. \ref{table:comparison}, protection performance is evaluated on four diffusion-based face swapping models. Across all settings, the baselines show limited protection capability. They frequently produce high ISM scores that indicate successful identity transfer. 
In contrast, we significantly reduce identity similarity for all target models. This effect is further consistently observed by the highest $L_2$ distances and lowest PSNR values, demonstrating that the swapped faces generated from our protected inputs deviate from source identity and exhibit more distortion.

Qualitative results are shown in Fig. \ref{fig:visual-2-intro} and Fig. \ref{fig:visual-main}. The baselines often fail to prevent identity transfer. In contrast, our adversarial faces result in either (i) severe degradation and artifacts, where the model yields irrelevant textures instead of key facial features, or (ii) a complete misinterpretation of the source identity, leading to the synthesis of identities unrelated to the source. These results confirm that by systematically disrupting the identity pathway, our method effectively prevents the accurate replication of the source identity.

\paragraph{Image quality.}
Tab. \ref{table:visual_quality} presents the quantitative assessment of the image quality of adversarial faces. Our method consistently outperforms baselines across all metrics. As visualized in Fig. \ref{fig:visual-2-intro}, the baselines tend to introduce visible noise and textural artifacts, particularly in smooth facial regions. In contrast, our method preserves fine details and maintains natural skin textures. This superior visual quality arises from performing adversarial search within the generative latent manifold, combined with a perceptual adaptive strategy. By selectively suppressing perturbations in sensitive regions, we keep the protective noise imperceptible to humans while retaining strong adversarial effectiveness.

\begin{table}[t]
    \centering
    \resizebox{0.9\columnwidth}{!}{
    \begin{tabular}{cccc ccc}
    \toprule
    Stage I & Stage II & \multicolumn{2}{c}{Stage III} & \multicolumn{3}{c}{Metric} \\
    \cmidrule(r){1-1} \cmidrule(lr){2-2} \cmidrule(lr){3-4} \cmidrule(l){5-7}
    $\mathcal{L}_{loc}$ & $\mathcal{L}_{id}$ & $\mathcal{L}_{attn}$ & $\mathcal{L}_{feat}$ & $L_2 \uparrow$ & ISM $\downarrow$ & PSNR $\downarrow$ \\
    \midrule
    \checkmark & & & & 0.0356 & 0.3537 & 35.0054 \\
    \checkmark & \checkmark & & & 0.0419 & 0.3073 & 32.2022 \\
    \checkmark & \checkmark & \checkmark & & 0.0448 & 0.2977 & 28.3642 \\
    \checkmark & \checkmark & \checkmark & \checkmark & 0.0454 & 0.2942 & 27.2620 \\
    \bottomrule
    \end{tabular}}
    \caption{Ablation study of each loss on the attack effectiveness.}
    \label{table:ablation_loss}
\end{table}
\begin{figure}[t]
    \centering
    \includegraphics[width=0.98\columnwidth]{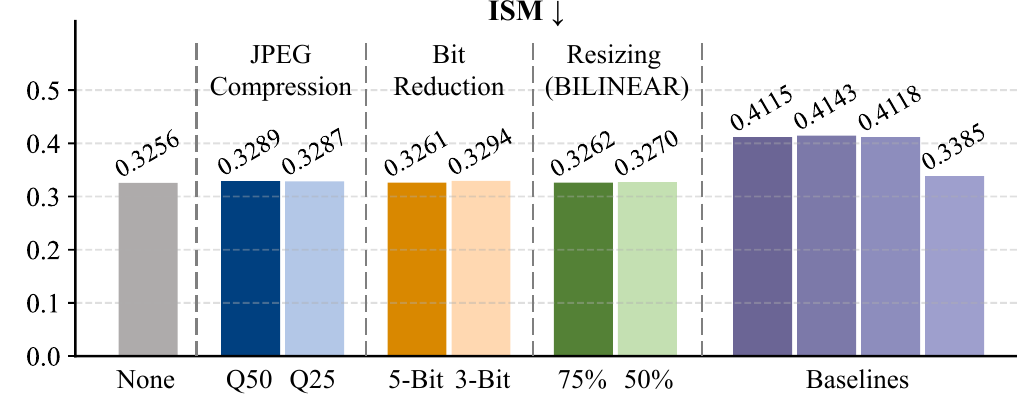}
    \caption{Robustness on various lossy operations. We report ISM on DiffFace on the CelebA-HQ dataset. The baseline methods from left to right are AdvDM, Mist, FACELOCK, and FaceShield.}
    \label{fig:visual-4-robust}
\end{figure}
\begin{figure}[t]
    \centering
    \includegraphics[width=\columnwidth]{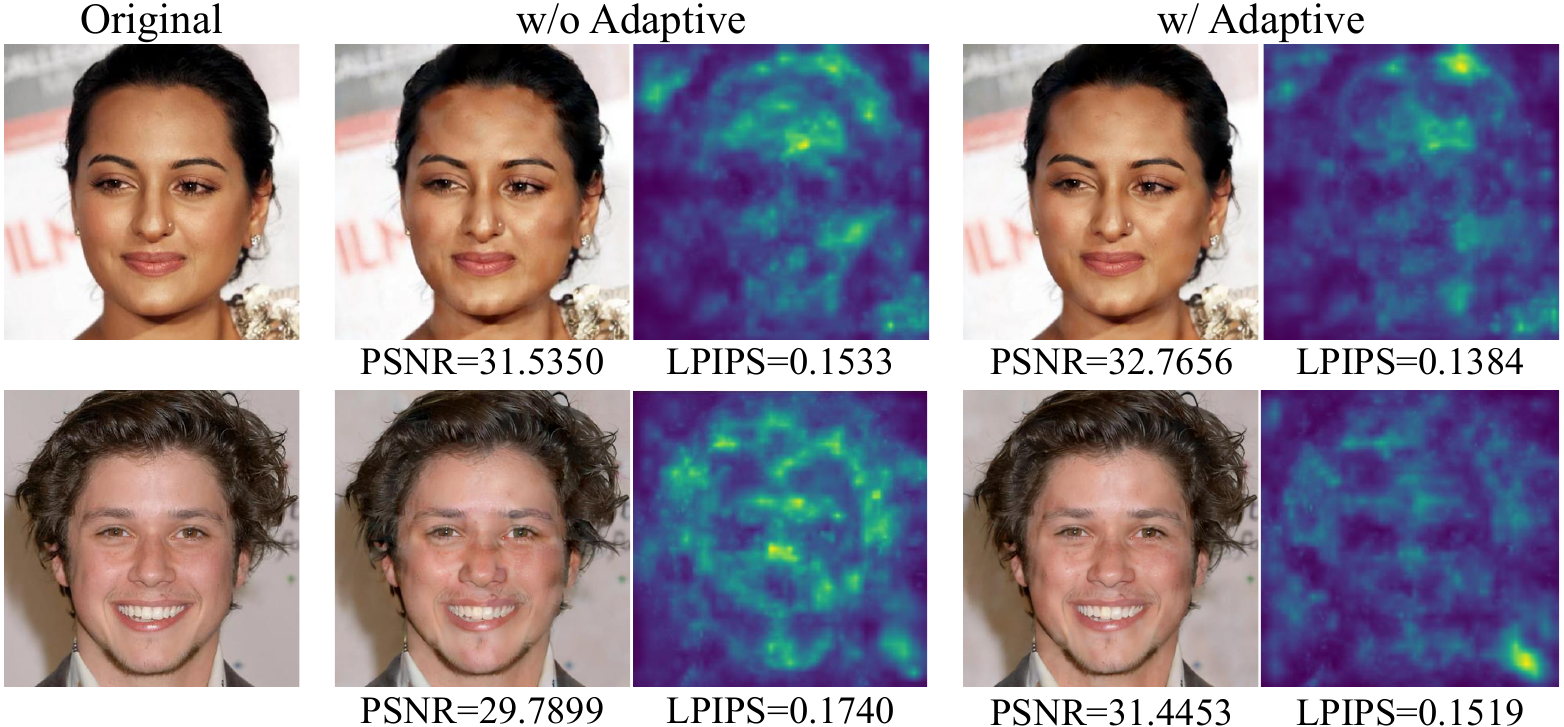}
    \caption{Ablation of perceptual adaptive strategy. We display protected images (left) and LPIPS distance maps (right). The brighter regions on the LPIPS distance maps indicate regions where subtle distortions are more noticeable.}
    \label{fig:visual-6-ablation_adaptive}
\end{figure}
\paragraph{User study.}
We conduct a user study with 50 participants. Participants assess two factors: (i) protection performance, and (ii) visual quality of the adversarial faces. As shown in Fig. \ref{fig:visual-5-user_study}, participants consistently rate our protection to be more effective and our adversarial faces to be more natural.

\subsection{Robustness}
We evaluate the robustness of VoidFace against common lossy operations, including JPEG compression, bit reduction, and resizing. As shown in Fig.~\ref{fig:visual-4-robust}, our method causes only minor performance degradation under these settings yet still outperforms the baselines. Our significant robustness stems from embedding rich adversarial content into the image structure via adversarial search within the generative manifold.

\subsection{Ablation Study}
\begin{figure}[t]
    \centering
    \includegraphics[width=\columnwidth]{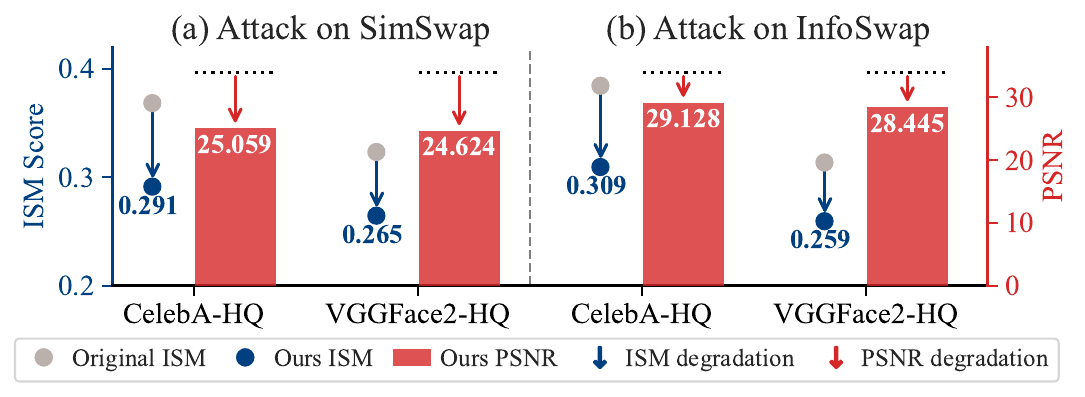}
    \caption{Transferability on two GAN-based face swapping models.}
    \label{fig:visual-7-transferability}
\end{figure}
\paragraph{Effect of each loss function.}
We validate the loss designed within our VoidFace. As detailed in Tab. ~\ref{table:ablation_loss}, each loss contributes to the overall defense. The physical disruption $\mathcal{L}_{loc}$ provides the destabilized baseline. Adding the semantic attack $\mathcal{L}_{id}$ leads to a substantial reduction in ISM. This confirms that perturbing the source representation in the embedding space effectively suppresses identity information. Introducing $\mathcal{L}_{attn}$ extends the attack into the generative domain. This results in significantly lower PSNR and higher $L_2$, demonstrating that interfering conditional injection disrupts the generation trajectory. Finally, incorporating $\mathcal{L}_{feat}$ further degrades the model's generative ability and yields the best performance. These results confirm that a systemic disruption across three domains is essential for defense.

\paragraph{Effect of perceptual adaptive strategy.}
We validate our perceptual adaptive strategy in Fig. \ref{fig:visual-6-ablation_adaptive}. When latent optimization is applied uniformly (w/o Adaptive), noticeable artifacts emerge in smooth facial regions. These artifacts correspond to high-intensity responses in the LPIPS distance maps. In contrast, our adaptive mechanism (w/ Adaptive) adjusts perturbation strength based on pixel-level perceptual feedback. This effectively reduces distortions in sensitive regions, yielding cleaner skin textures and superior evaluation metrics.

\subsection{Transferability}
\label{sec: transferability}
To demonstrate our applicability, we further conduct experiments on two additional GAN-based face swapping models, SimSwap~\cite{simswap} and InfoSwap~\cite{infoswap}. As shown in Fig. \ref{fig:visual-main} and Fig. \ref{fig:visual-7-transferability}, our method leads to degradation in model performance, demonstrating effective transferability to GAN architectures.
\section{Conclusion}
This paper presents VoidFace, a systemic defense method against diffusion-based face swapping. The method leverages the sequential structure of the swapping pipeline and applies a cascading pathway disruption strategy, targeting key bottlenecks across physical, semantic, and generative domains. VoidFace further adopts perceptually adaptive optimization in the latent manifold to balance attack potency and visual quality. Extensive experiments show that VoidFace consistently outperforms existing methods. Overall, this work offers a promising direction for safeguarding facial identity.






\bibliographystyle{named}
\bibliography{ijcai26}

\end{document}